\documentclass[letterpaper, 10 pt, conference]{ieeeconf}  

\IEEEoverridecommandlockouts                              

\usepackage{graphicx} 
\graphicspath{ {./images/} }

\usepackage[rightcaption]{sidecap}

\usepackage{wrapfig}
\usepackage{gensymb}

\usepackage{siunitx}
\usepackage{xcolor}
\usepackage{float}

\usepackage{multirow}
\usepackage{tabu}
\usepackage{subcaption}
\usepackage{amssymb}
\usepackage{balance}

\title{\LARGE \bf
Fast Perception, Planning, and Execution for a \\ Robotic Butler: Wheeled Humanoid M-Hubo
}

\author{
Moonyoung Lee$^{1}$, Yujin Heo$^{1}$, Jinyong Park$^{1}$, Hyun-Dae Yang$^{1}$\\
Ho-Deok Jang$^{2,\dagger}$, Philipp Benz$^{2,\dagger}$, Hyunsub Park$^{1}$, In So Kweon$^{2}$, Jun-Ho Oh$^{1}$
\thanks{$^{1}$Is with the Humanoid Robot Research Center, Department of
Mechanical Engineering, Korea Advanced Institute of Science and Technology, 291 Daehak-ro, Yuseong-gu, Daejeon 34141, Korea 
        {\tt\small jhoh@kaist.ac.kr}}%
\thanks{$^{2}$Is with the Robotics and Computer Vision Laboratory that is in charge of object perception part, Department of Electrical Engineering, Korea Advanced Institute of Science and Technology, 291 Daehak-ro, Yuseong-gu, Daejeon 34141, Korea. E-mail: {\tt\small\{hdjang, pbenz, iskweon77\}@kaist.ac.kr}. $\dagger$ denotes the equal contribution for the part.}%
}

\begin{document}

\maketitle
\thispagestyle{empty}
\pagestyle{empty}

\begin{abstract}

As the aging population grows at a rapid rate, there is an ever growing need for service robot platforms that can provide daily assistance at practical speed with reliable performance.
In order to assist with daily tasks such as fetching a beverage, a service robot must be able to perceive its environment and generate corresponding motion trajectories.  
This becomes a challenging and computationally complex problem when the environment is unknown and thus the path planner must sample numerous trajectories that often are sub-optimal, extending the execution time. 
To address this issue, we propose a unique strategy of integrating a 3D object detection pipeline with a kinematically optimal manipulation planner to significantly increase speed performance at run-time.
In addition, we develop a new robotic butler system for a wheeled humanoid that is capable of fetching requested objects at $24\%$ of the speed a human needs to fulfill the same task. The proposed system was evaluated and demonstrated in a real-world environment setup as well as in public exhibition.

\end{abstract}

\section{INTRODUCTION}

Aging of population is an increasing problem in modern society.
The old-age dependency ratio, ratio of elderly above 65 per 100 people between 15 and 64, is as high as 45\% in Japan \cite{oldage}. Shortage of available human assistants to tend to elderly’s needs has led to their deterioration in quality of life. From this need, service robots with manipulators that can assist people in daily living such as fetching a drink has gained wide interest.

A typical mobile service robot involves multiple key components such as object detection, trajectory planning, grasping, localization and mapping, navigation, and motion control. Each of these subtasks are challenging and constitute an active area of research on their own. However, assembling and ensuring the tight interplay between these subtasks into a working system to solve the overall task is even more important.

\begin{figure}[h]
    \centering
    \includegraphics[scale=0.7]{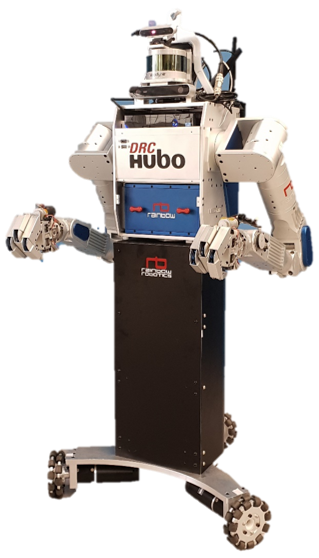}
    \caption{Service Robot Platform, M-Hubo}
    \label{fig:M-Hubo}
\end{figure}

Current state-of-the-art service robots solving the task of fetching a requested object are rather slow in their execution speed, making them impractical for deployment in the end-user stage.
Manipulation tasks in particular are found to be slow among service robots due to the consecutive nature of their perception and planning modules, while the robot is completely stationary throughout this process.
This strategy is primarily selected to increase accuracy in modeling the surrounding world as well as performing random-sampled-based planning to ensure grasping without collisions. Although this approach is appropriate in situations where the environment is not mapped or highly dynamic, the overall execution time to complete perception and manipulation can be long. Furthermore, replanning paths from dense perception information becomes redundant in static environments, where a map is already available, especially with accurate localization information.

 To address this issue, we propose a unique strategy of integrating a 3D object detection pipeline with a kinematically optimal manipulation planner to significantly reduce the overall manipulation execution time. 
This integration enables fast pose estimation of requested objects, while smooth trajectories are generated in under $50$\si{milliseconds} in run-time by the planner.

We develop a new, fully-autonomous robotic butler system for a wheeled humanoid. In this work, we focus on one particular application for service robot: fetching and serving drinks at comparable human-like speeds in a static indoor environment. 

We provide an in depth description of our hardware platform as well as for each module that enables the drink fetching service. We further evaluate our system on the component level in the areas of object perception, path planning, and localization. Finally, for the overall system is evaluated and compared to previous methods in a real world environment setup.

Our main contributions are
\begin{itemize}
    \item Integrated service robot system for fetching and serving drinks at 24\% of human-like speed, which is  faster than state-of-the-art mobile manipulator platforms. 
    \item Evaluation and demonstration of the proposed system in a real world environment setup as well as in a public exhibition  
    
\end{itemize}
The remaining of the paper is structured as follows. 
We review previous works and compare them briefly to our system in section II.
Section III details the robot hardware platform and software architecture developed for handling high-level tasks with real-time motion controllers. 
Section IV discusses the object perception process.
Section V discusses concise trajectory planning,
and Section VI discusses accurate localization for small indoor spaces. 
We conclude with results and discussion in Section VII. 

\section{RELATED WORK}
In the last years, several works had been proposed describing service robots with manipulators.
Care-O-Bot 3 \cite{careobot} is a mobile robotic system with one 7 DoF arm and a tray for serving drinks. The authors report an overall success rate of $40\%$, even though individual subsystems perform much better. This performance drop indicated the importance of system integration. 
HERB~\cite{herb} is also mobile manipulator with one 7 DOF arm on a segway mobile base that perform service tasks in kitchen settings. Although HERB shows high a success rate, Srinivasa et al.~\cite{herb} report task execution times to be slow slow, with 55 seconds required to navigate and re-localize, and 30 seconds to pick up a cup while the robot is stationary.
The task of fetching a drink closely resembles the one in this paper, whereby our robot platform utilizes two manipulators in contrast to one in \cite{careobot, pr2smach}, adding complexity, but also increasing efficiency of our robot.
The authors of \cite{pr2smach} proposed a service robot fetching a drink from a refrigerator, by deploying the PR2 robot platform. From their 30 trials, Bohren et al. remarked on high occurrence of failed detection resulted in failure recovery procedure, prolonging execution time of approximately 110 seconds per fetch. 

The next related works highlight their tight integrated perception and manipulation tasks on a wheeled humanoid, similar to our robot platform. 
Work in \cite{rollinjustin} shows impressive dexterous manipulation by catching flying balls under 1 second with 80\% success rate with Rollin Justin platorm.
Uniquely similar to our platform, Rollin Justin also utilizes real-time based motion controller for improving manipulation precision. However, focus in both \cite{rollinjustin, armar} are on integrated manipulation tasks stationed in fixed space rather than navigating distances to fetch objects.
 Pyo et al. \cite{servicerobot} use a two armed wheeled humanoid for application of service robot to assist elderly by fetching requested beverage in a real world environment setup with similar dimension to ours. Their service robot, however, utilized external environmental sensors ,resulting in an informationally structured environment, and focused on more robust collision-free planners at the cost of execution time. Their reported execution time per fetched drink was 312 seconds. 

All previous works discussed here have in common that they lack fast execution times. In this work we address this problem, by proposing a mobile robot platform, capable of fetching drinks faster than previously proposed approaches for static environments.

\section{SYSTEM SETUP}

\subsection{Robot Platform Hardware }

\begin{figure}[t]
    \centering
    \includegraphics[scale=0.5]{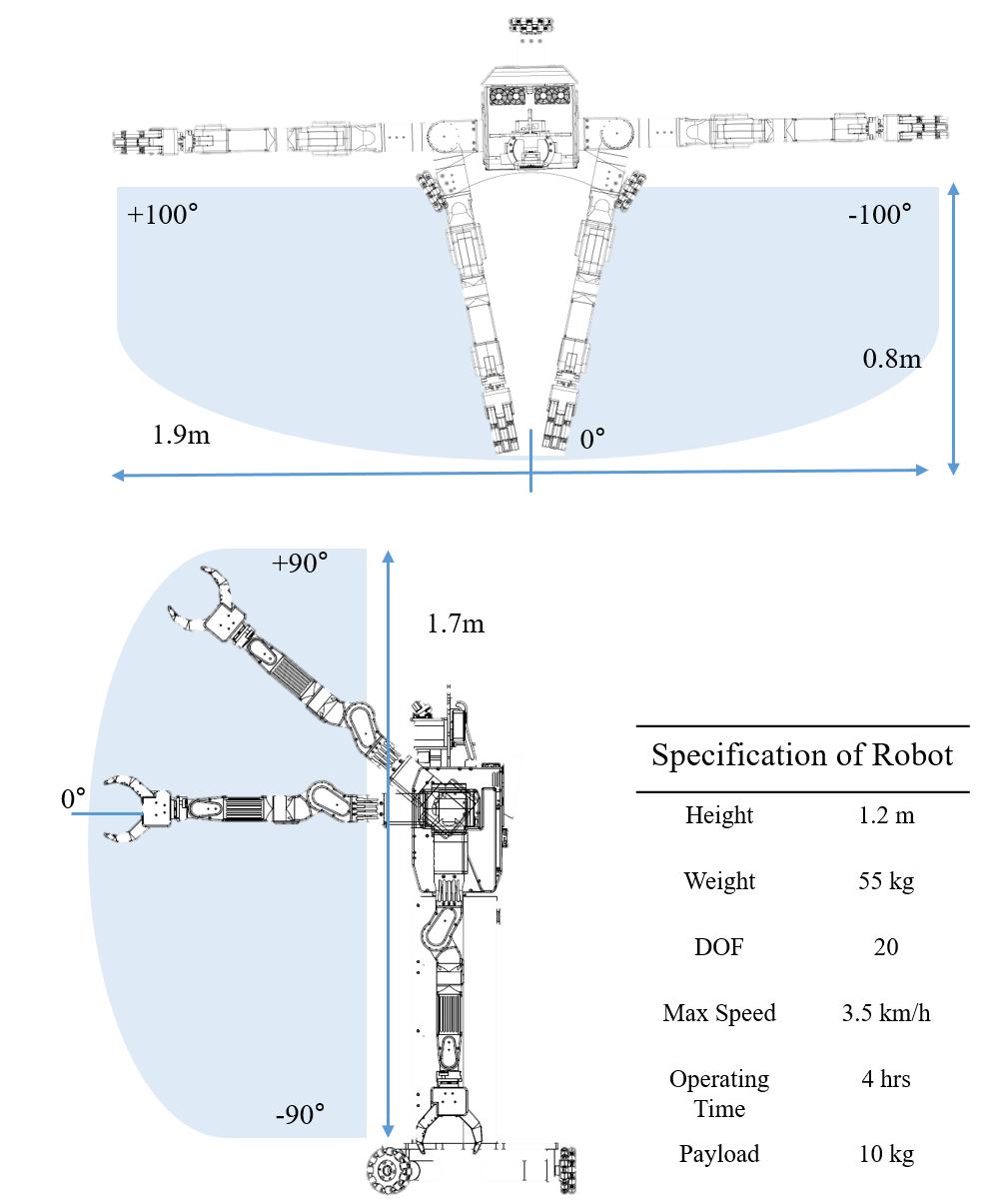}
    \caption{Workspace and overall robot specification}
    \label{fig:workspace}
\end{figure}

The robotic butler Hubo is a humanoid robot with an omni-directional base. The hardware setup consists of 20 degrees of freedom total. It contains two 7 DOF manipulator arms, which have optimized joint limits to provide maximum workspace of $100$\degree each with $0.8$ m reach as shown in Figure~\ref{fig:workspace}. The wide workspace of combining left and right manipulator, in addition to the omni-directional base control, reduces the need for re-orientating the robot before object grasping, which hastens the fetching service. The robot arm is equipped with shape adaptive 3 finger grippers rated for $2$ kg payload at the finger and $10$ kg payload at the arm. The omni-directional base, which can reach a maximum speed of 3.5 km/h, with the extra DOF in the robot waist, provide effective locomotion control to maintain vision sensors' field of view on the requested beverage. This hardware setup to maintain the field of view on the region of interest reduces the fetching time by enabling to detect shortest navigation paths to an object in appearance of an obstacle. In order to prevent undesirable tipping in robot's pitch during high velocity  manipulation or base motion, the robot is bottom heavy, including the Ni-ion battery packs (48V, 11Ah).   

The authors utilize the manipulator design from the prior DRC-HUBO+ humanoid robot ~\cite{jungDRC}, which enables precise position control due to high rigidity in manipulator design and minimal jitter in real-time joint reference communication. Precise position control is crucial for our fast service application in order to reduce the probability of grasp failure recovery as well as external collision due to perception inaccuracy.

\subsection{Software Architecture }

The software architecture is intuitively divided among perception and motion control. For subtasks related to perception, we dedicate a separate embedded Vision PC operating on Robot Operating System (ROS) middleware. For motion control, we dedicate a separate embedded motion PC operating on the custom PODO software framework~\cite{jungDRC} to leverage real-time communication of motor control. 

\begin{figure}[t]
    \centering
    \includegraphics[scale=0.4]{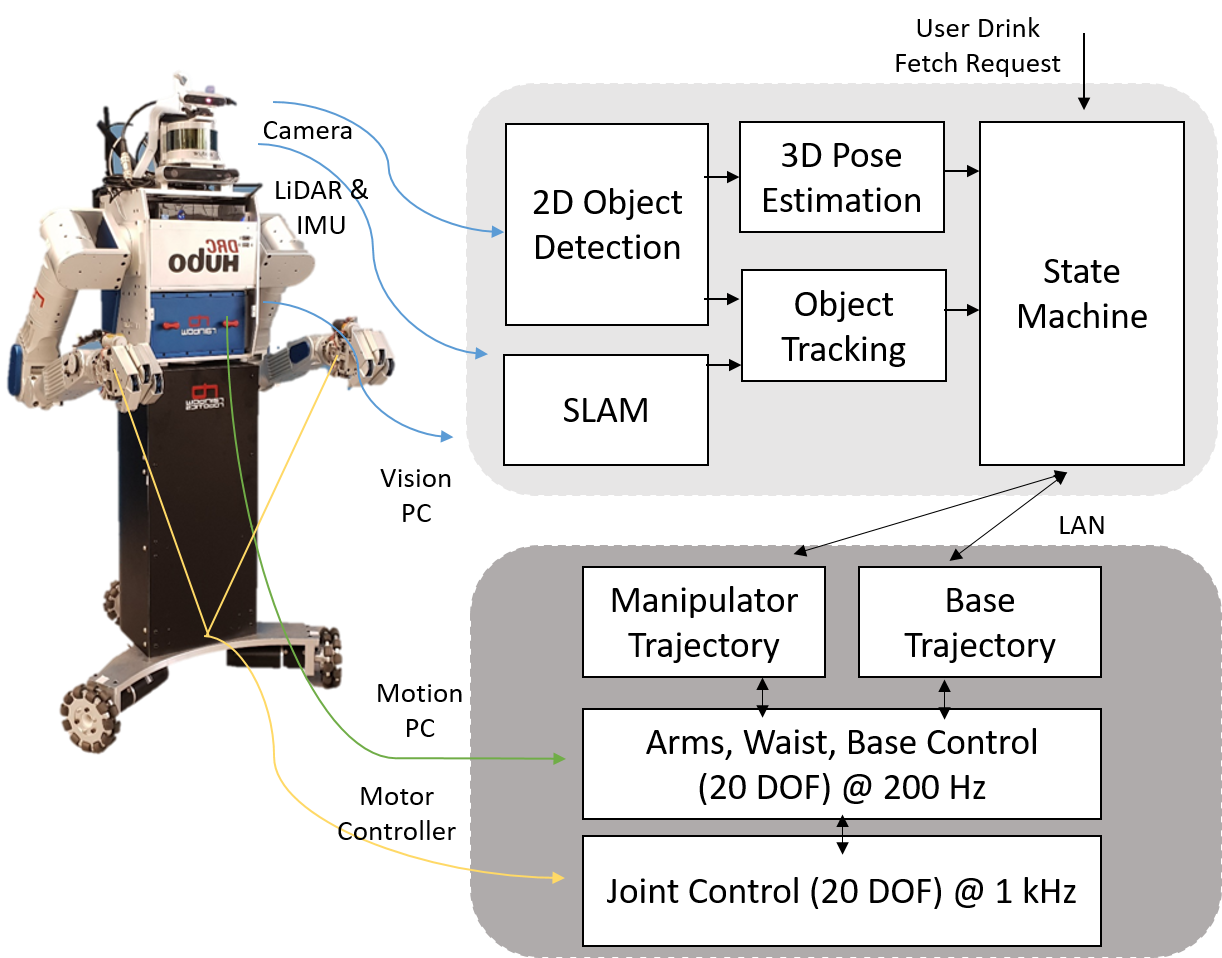}
    \caption{Overall architecture for service robot that integrates ROS for high-level perception and PODO for low-level real-time motion}
    \label{fig:software}
\end{figure}

Within the Vision PC exist modules that model the environment and relevant objects using the robot's vision sensor suite (detailed in section IV). The modules include 2D Object Detection, 3D Pose Estimation, Object Tracking, SLAM, and a finite state machine. With the processed environment or object data, the Vision PC transmits tasks commands and associated processed data to the Motion PC. For our state machine, we leverage the task-level executive system SMACH ~\cite{pr2smach} to define the states used for the fetch drink service, which is described in Figure~\ref{fig:time lapse}. For the implemented simple scenario, only a FSM with total of seven states was required. The state transition conditions were determined by the flags indicating robot's active state for the corresponding motion or success of object detection requests. 

The Motion PC translates the high level task to generate manipulator or base trajectories, which then generate joint configuration by solving the inverse kinematic analytically, which are then executed through motion controllers. In addition, the Motion PC continuously updates the Vision PCs the task execution state, current robot state information, which are joint encoder and F/T sensor data. 
The PODO software framework~\cite{jungDRC} allows precise motion control through reduction of communication jitter. This RT framework ensures accuracy of joint information and sensor updates by synchronizing multiple threads at very regular intervals of 200 Hz to the Motion PC through a CAN bus interface. The real-time Linux interface is based on Xenomai RT.

The software architecture is depicted in Figure~\ref{fig:software}. 
The models for the Vision PC and Motion PCcomputers are Alienware ASM201 with i7 processor, 16GM RAM, GTX 960 GPU, and Intel NUK6i7KYK with i7 processor, 8GB RAM, respectively.


\newcommand{\figref}[1]{Figure.~\ref{#1}}
\newcommand{\tabref}[1]{Table~\ref{#1}}
\newcommand{\eqnref}[1]{Eqn.~(\ref{#1})}
\newcommand{\secref}[1]{Sec.\ref{#1}}
\newcommand{\ie}{{\it i.e.}}
\newcommand{\eg}{{\it e.g.}}
\renewcommand{\vec}[1]{\mathbf{#1}}


\section{OBJECT PERCEPTION}


\begin{figure}[t]
    \centering
    \vspace{0.5cm}
    \includegraphics[width=\linewidth]{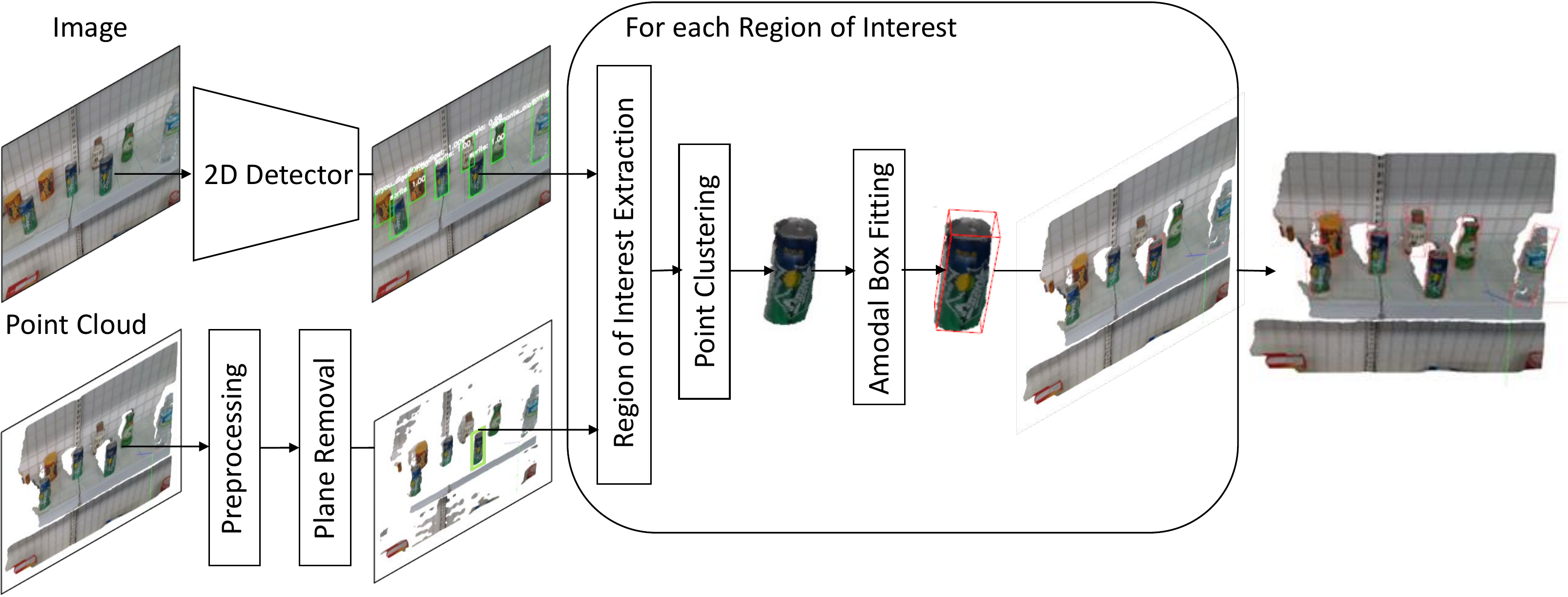}
    \caption{\textbf{3D object detection pipeline.} We cascade a deep neural network based 2D object detector and a conventional 3D box fitting algorithm to perform the 3D detection task. This configuration provides both \textit{high recall} for small objects and \textit{scalability} in building a training dataset for the 3D detection task.}
    \label{fig:detection_pipeline}
\end{figure}

\begin{table}[!t]
\centering
    \resizebox{1.0\linewidth}{!}{
    \begin{tabular}{l|c|c|c|c|c|c}
    \hline
    \multirow{3}{*}{model} &  \multirow{3}{*}{input size} & \multicolumn{5}{c}{dataset}  \\
    \cline{3-7}
                           & &\multicolumn{3}{c|}{VOC 2007 \texttt{test}}  &  \multicolumn{2}{c}{M-HUBO \texttt{test}}\\
    \cline{3-5}\cline{6-7}
                           & &   fps  & params         & mAP           & params          & mAP           \\
    \hline
    \hline
    YOLOv2~\cite{redmon2016yolo9000} & 416$\times$416  & 67   & -              & 76.8          & -               & -   \\
    SSD~\cite{liu2016ssd}            & 300$\times$300  & 46   & 26.5M          & 77.2          & -               & -   \\
    StairNet~\cite{woo2017stairnet}  & 300$\times$300  & 30   & 28.3M          & 78.8          & 28.2M           & 92.6\\
    PASSD~\cite{jang2018passd}       & 320$\times$320  & 50   & \textbf{24.9M} & \textbf{81.0} & \textbf{24.8M}  & \textbf{96.6} \\
    \hline
    \end{tabular}
    }
    \caption{\textbf{2D Object detection results.} All models are evaluated on PASCAL VOC 2007 \texttt{test} set and M-HUBO \texttt{test} set}
\label{tab:2d_detection}
\end{table}

\begin{figure}[t]
    \centering
    \includegraphics[scale=0.4]{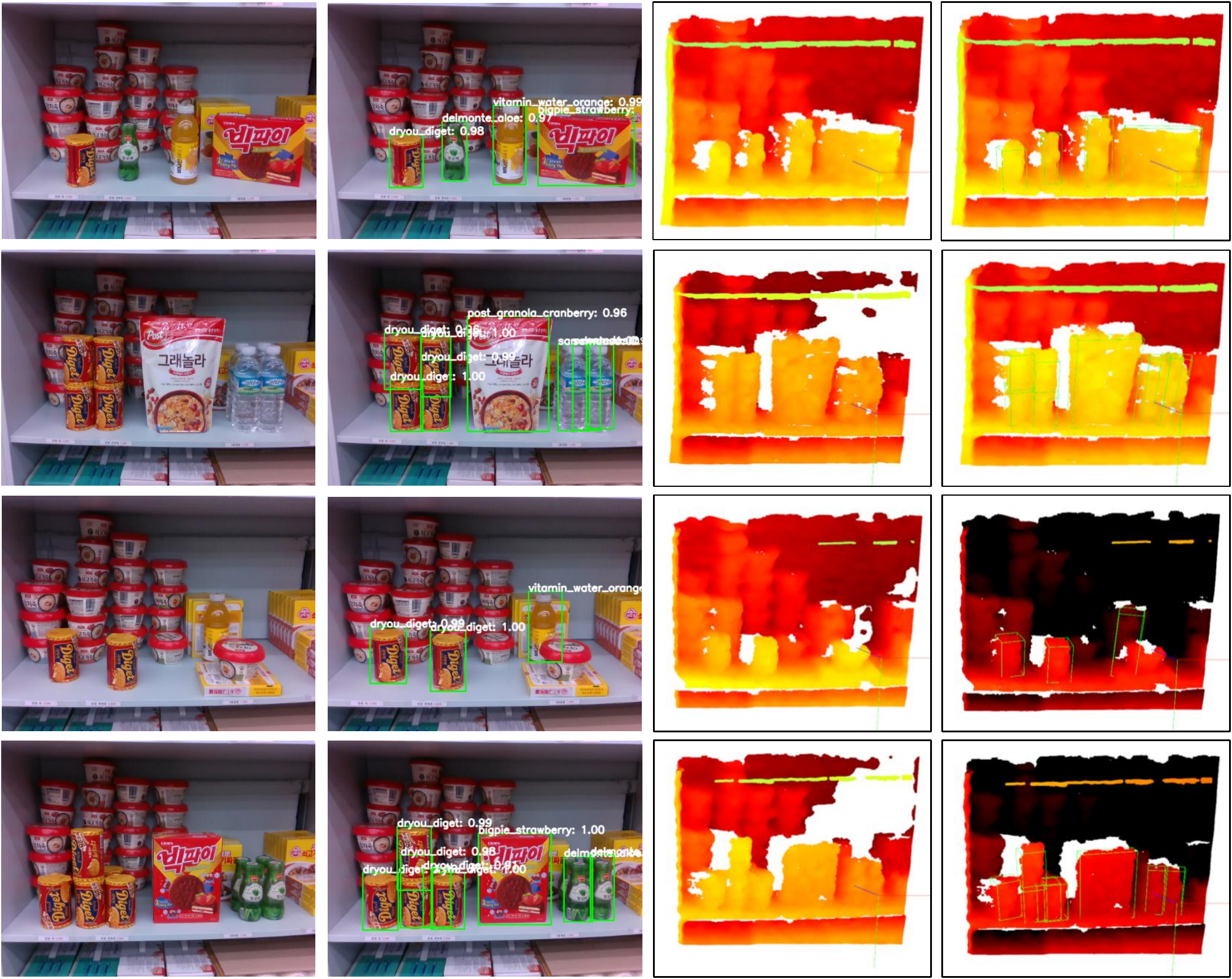}
    \caption{Qualitative results of the perception algorithm captured in a real convenience store. Each row captures a different scenario. The four columns depict from left to right the input image, 2D detection results, input point cloud and box fitting result. Bounding boxes are depicted in green for 2D and 3D case. }
    \label{fig:detection_pipeline_qualitative_results}
\end{figure}

\subsection{Head Sensor Pack}
The head sensor pack of M-Hubo consists of two RGB-D cameras (Realsense D415, ZED stereo camera) and one LiDAR (Velodyne Puck VLP16) as shown in Figure~\ref{fig:M-Hubo}. The RGB-D cameras serve the purpose of providing RGB images as well as depth information of the nearby environment. LiDAR is used to capture the overall and long-range surrounding environment. To align the field of view (FOV) of Realsense D415 to the region of interest of the robot, we tilted the camera by an angle of \ang{30} along vertical dimension.


\subsection{Perception Pipeline}

For robot manipulation, the 3D location and extent of the objects of interest are required, which can be provided by a 3D object detection algorithm. With the arise of deep learning, 3D object detection algorithms have also been greatly benefited from this new technology~\cite{song2016deepsliding}. However, most of these algorithms work in a supervised manner, where annotations for the target task should be provided. While supervised setting generally performs better than semi-supervised or unsupervised approaches, the cost for annotating in the 3D space is non-trivial. Even though the annotation cost is affordable in 2D space to some extent, extending it to 3D space is nearly prohibitive due to its high costs.  In other words, annotating in 3D space highly constrains the \textit{scalability} (e.g.,  category) in building training dataset for deep neural network.

To mitigate the problem, we design a \textit{scalable} 3D object detector by combining a mature deep neural network based 2D object detector with a conventional 3D box fitting algorithm.
The 2D detector first proposes regions of interest (RoI) for target objects using RGB input image. Then, the subsequent box fitting algorithm estimates the 3D bounding box of the objects using depth given in each RoI. By lowering the level of annotation from 3D to 2D space, the cost of annotation is substantially reduced, resulting in \textit{scalable} 3D object detection algorithm. In addition, the algorithm is benefited from the high recall for small objects by the deep neural network based 2D detector. The overall pipeline is shown in Figure~\ref{fig:detection_pipeline}, and the details will be described below.

For the instantiation of 3D object detector, we adopt PASSD~\cite{jang2018passd} as a 2D object detector, which ensures both real-time speed and high accuracy, making it suitable for our robot environment. In brief, the PASSD~\cite{jang2018passd} improves the detection accuracy of one-stage detectors~\cite{liu2016ssd} while maintaining its high efficiency by incorporating the \textit{propose-and-attend} mechanism of two-stage detectors~\cite{ren2015faster} in an efficient manner,~\ie~integration of region proposal generation and region-wise classification in a single-shot manner without stage division. The 2D detector provides RoIs of the target objects. Then, we perform region-wise amodal 3D box fitting following preprocessing and plane removal. In the preprocessing step, we filter out 3D points outside of a certain range ($1.5$ m in this work). For the plane removal, we use RANSAC to segment the largest plane (\ie, floor plane), where the plane-normal is kept for later box fitting operations. For the region-wise 3D box fitting, we extract the 3D points in the RoI, and apply euclidean clustering to obtain clean 3D points of the target object. Then we project the clustered 3D points to the floor plane, and fit the rectangle around them. The final 3D bounding box is obtained by extending the fitted rectangle along the normal of the floor plane by the known height of the object. We represent the final 3D bounding box as its eight corner points and a center point of the box.

\subsection{Perception Evaluation}
To validate our perception system on the target task, we generated a dataset which resembles the real scenario of our target task. The dataset is composed of total 1757 images with 1380, 280 and 177 images for train, validation and test respectively. The dataset covers 13 object categories that are commonly observed in convenience stores such as beverages or snack boxes. The objects in the image are annotated with 2D bounding boxes and their category labels. 

For the 2D detector~\cite{jang2018passd}, we fine-tune the 2D detector on the M-HUBO dataset after pre-training on a large-scale dataset (\ie, PASCAL VOC~\cite{everingham2010pascal}) to resolve the data scarcity and ensure generalization capacity. We compare our 2D detector,~\ie, PASSD~\cite{jang2018passd}, with other strong candidates which also have real-time capability. As shown in~\tabref{tab:2d_detection}, the deployed algorithm~\cite{jang2018passd} outperforms not only on the large-scale dataset (\ie, PASCAL VOC~\cite{everingham2010pascal}), but also on the custom M-HUBO dataset by large margin and with less parameters. It also runs in real-time.

For the 3D bounding box, we evaluate empirically by comparing estimated positions from the 3D object detector with measured object positions relative to the camera since there is no ground-truth available for 3D objects. We conduct this process for $3$ different objects in $10$ different positions, and found that the 3D object detection pipeline performs within an accuracy of $1$ cm. This accuracy is sufficient for the target task since small inaccuracies in the 3D box positioning can be compensated through the following grasping mechanism.


Moreover, we further evaluate our whole perception system under a real convenience store environment. The real-world scenario is challenging due to its cluttered scene such as grouped or stacked objects and objects with partial occlusion. The qualitative results in~\figref{fig:detection_pipeline_qualitative_results} show that our system in overall can handle this setup well with marginal limitations for the densely grouped objects. The issue might arise from the small training samples for such cases, and can be mitigated with more training samples.



\section{PATH PLANNING}
\subsection{Base Path Planning}
Upon receiving a new fetch drink request, the robot has to determine its base path in a 2D space and omniwheel velocities to follow that trajectory. A popular applied solution to base path planning, especially for navigation in 2D grid space, includes heuristic search A*. Although the authors initially applied dynamic replanning based on the A* algorithm, we observed reduced average base velocities due to constant changing of reference velocity during the robot's motion. 

Given that our application focuses on a static indoor environment, where the environment is already mapped prior to service exeuction (discussed in SectionVI) with rough landmark positions in the scene, we adopt a simpler navigation scheme for the gain of maximizing average velocity during the fetch task. We first implement a navigation scheme in which the general location of the end goal is known prior to fetching, and creates a $1-cosine$ trajectory in space with a trapezoidal velocity profile to maximize average velocity. Upon arriving at the general location, the robot then scans for the object before grasping it.              

We further reduce the fetch execution time with our second navigation scheme by utilizing the above 2D detection algorithm to track the object with respect to the robot as it is approaching it. 
Given the estimated object pose $y$, the reference velocity $V_y$ is scaled linearly to align itself with the tracking object using the equation below.

 \begin{equation} 
V_y =  \frac{ ( (Vy_{max} - Vy_{min}) * (y_{obj} - y_{min}) )} {(y_{max} - y_{min})}
\end{equation}

$Vy_{min}$ and $Vy_{max}$ are constraint so that the omniwheel controller maintains maximum average velocity. 
$y_{min}$ and $y_{max}$ are constraint so that upon arriving at the tracked object, the robot's left or right manipulator is positioned with a desired offset from the objects which provides a better kinematic solution for grasping. This offset fixed so that the robot's manipulator and object are aligned along the y-axis upon arriving in front of the object, avoiding much of the configurations that result in singularities if the object were directly in-front of the robot's center rather than the manipulator. Once the robot comes to a halt in-front of the object, the 3D perception described above is separately requested leading to manipulation planning. 


\subsection{Manipulation Path Planning }

Given the estimated object pose with respect to the robot position while approaching the target object, the path planner has to determine where in space to place the end-effector and its corresponding joint configurations. Typical constraints obeyed in order to formulate a path can be joint range limits, joint velocity limits, or geometrical constraints to prevent collisions internally or externally. 
As the number of DOF to consider increases in path planning, the more complex and computationally heavy the problem becomes. 
 To simplify the problem, we strategically exclude the base and waist joints in the planning process knowing that the robots base position and orientation upon arrival with respect to the object are fixed through the base path planning discussed above. In addition, we can make further simplifications by asserting that only the closest manipulator (left or right) to the object upon arrival will be used for grasping. 

\begin{figure}[t]
    \centering
    \includegraphics[scale=0.8]{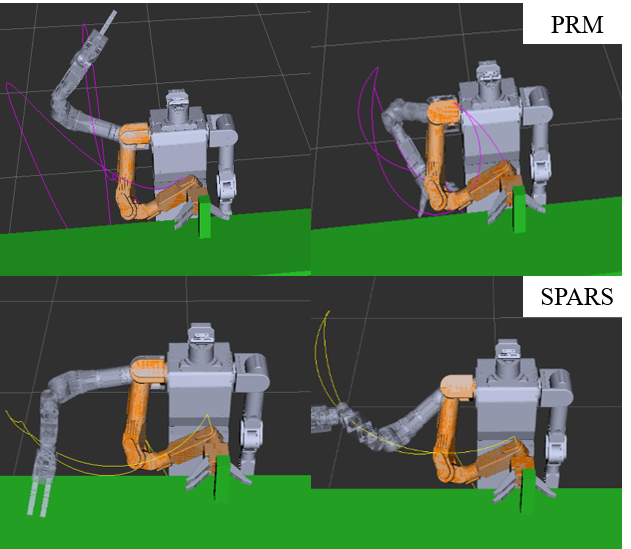}
    \caption{Cases of sub-optimal trajectory generated for the robot's end effector when utilizing sample based planners}
    \label{fig:path_planned}
\end{figure}

\begin{figure*}[t!]
    \centering
    \includegraphics[scale=0.4]{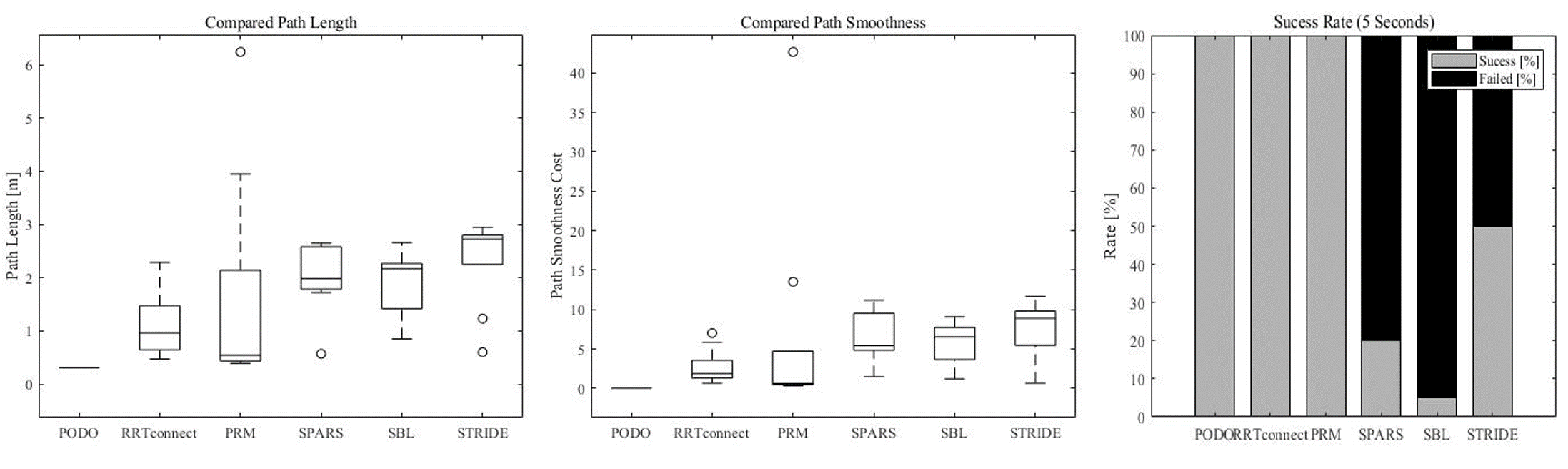}
    \caption{Evaluation of the proposed planner against popular sample-based path planners. Evaluation criteria are path length, smoothness, and plan success rate}
    \label{fig:path_graph}
\end{figure*}
We adapt the kinematically optimal planner from \cite{jungDRC} to create smooth-like manipulator trajectory by adhering to joint range limits and joint velocity limits. In comparison to well-established sample based planners such as RRT and PRM~\cite{PRM}
, our approach always and directly generates smooth trajectories at much faster rates at the cost of neglecting collision checking at run-time.  
While mentioned sample based planners excel in geometrically complex problems to solve for collision-free trajectories, they are not appropriate for our application because in addition to long computation time they neither guarantee path generation nor path optimality, which can result in service execution failure.     

Instead, to reduce planning time given sufficient prior knowledge of the static environment and object dimensions, we  manually fine-tune the generated trajectory through simulation trials and utilize those parameters at run-time to prevent collisions. Upon evaluating our kinematically optimal planner against traditional random sampling based planner (RRTconnect, PRM, SPARS, SBL, STRIDE), we observe that our path planner is well applied.

We utilize the MoveIt! Platform with the OMPL library~\cite{moveit}
to apply these planners on our robot's manipulator to move from default position to the position just prior to grasping an object. We allow a window of $5000$ ms for the planner to optimize and evaluate the generated trajectory based on 3 factors: path length, path smoothness, and plan success rate. 
Path length and smoothness are calculated from measuring the end-effectors pose $q$ over time and using the equations, which are established in~\cite{equationpath, equationsmooth}.

 \begin{equation} 
\text{Manhattan Distance} =  \sum_{k=2}^{N} d (q_k - q_{k-1})
\end{equation}

 \begin{equation} 
\text{Smoothness Cost}[q] =  \frac{1}{2} \int_{0}^{1}  || {\frac{d}{dt} q(t)} ||^2 dt
\end{equation}

As shown in Figure~\ref{fig:path_graph}, trajectory generated from PODO is always a fixed path of $0.4$ m, while other planners can vary in a plan success rate in addition to yielding longer trajectories. 
In static environments and for applications that require fast run-time trajectory planners, our kinematically optimal planner outperforms the random sample based planner. However, 
in order to prevent collisions with our planner, sufficient prior geometric knowledge of the environment must be obtained through mapping and require path verification through simulation before run-time.

\section{LOCALIZATION}

Localizing provides the robots base pose with respect to global the coordinate frame of the environment. For fast service robot applications in an indoor environment, we focused on run-time accuracy rather than optimizing the post-processed map. 
This is a significant contribution from our previous DRC-HUBO+ platform~\cite{jungDRC}, which had no visual localization ability and relied on human tele-operation for distanced motions. 

One challenge is maintaining localization accuracy without the aid of visual landmarks such as QR codes when the robot is confined to small room space. Our environment setup is a $10$ m $\times$ $8$ m room with homogeneous patterned walls.

The robot can utilize two sensor configutations for localization: LiDAR and IMU based, and RGB-D based. For LiDAR and IMU based localizaiton, we apply the Google Cartographer package in ROS, which 
runs frontend threads to create a succession of submaps and backend threads to match created submaps for global loop closures~\cite{cartographer}. We fine-tune the package for increased localization accuracy for the mock setup environment by modifying parameters for scan range, optimization resolution, period, and weight parameters for odometry.

For RGB-D based localization we apply Real-Time Appearance Based mapping, which is a graph-based SLAM using an incremental SIFT based loop closure detector~\cite{rtabmap}.
Since the dense point cloud from RGB-D camera $1$ imposes heavy computation for the vision PC embedded in the robot, we opt for a low resolution of $640 \times 480$ larger than $30$ FPS.

\begin{figure}[t]
    \centering
    \includegraphics[scale=0.65]{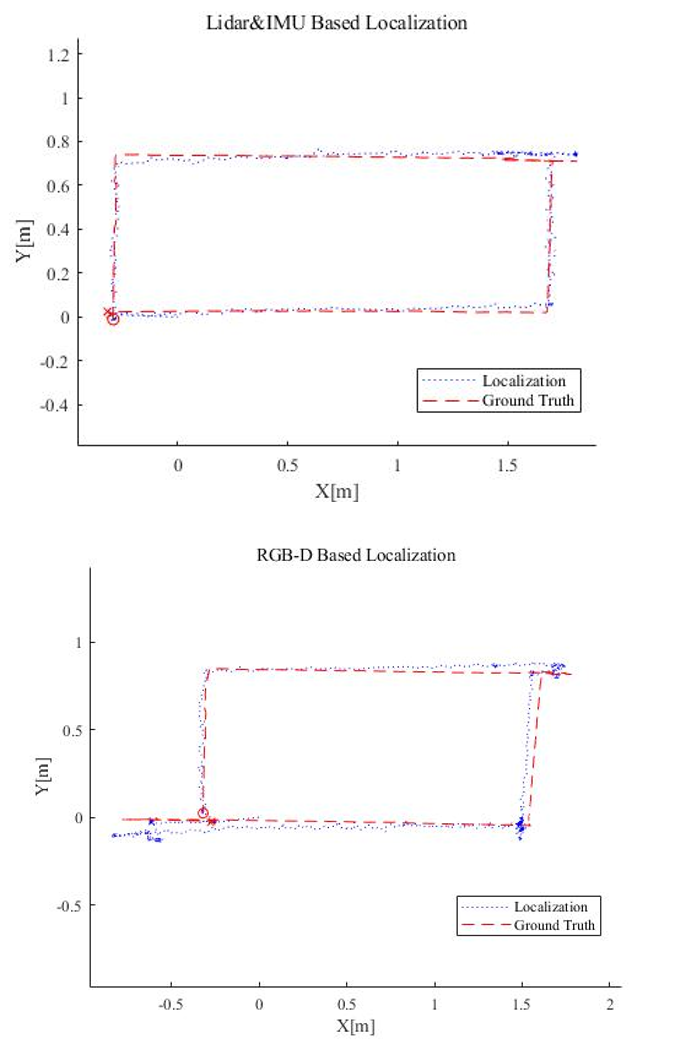}
    \caption{Estimated localization and ground truth data as robot moves throughout static indoor room. }
    \label{fig:slam}
\end{figure}

\begin{figure*}[t!]
    \centering
    \includegraphics[scale=0.7]{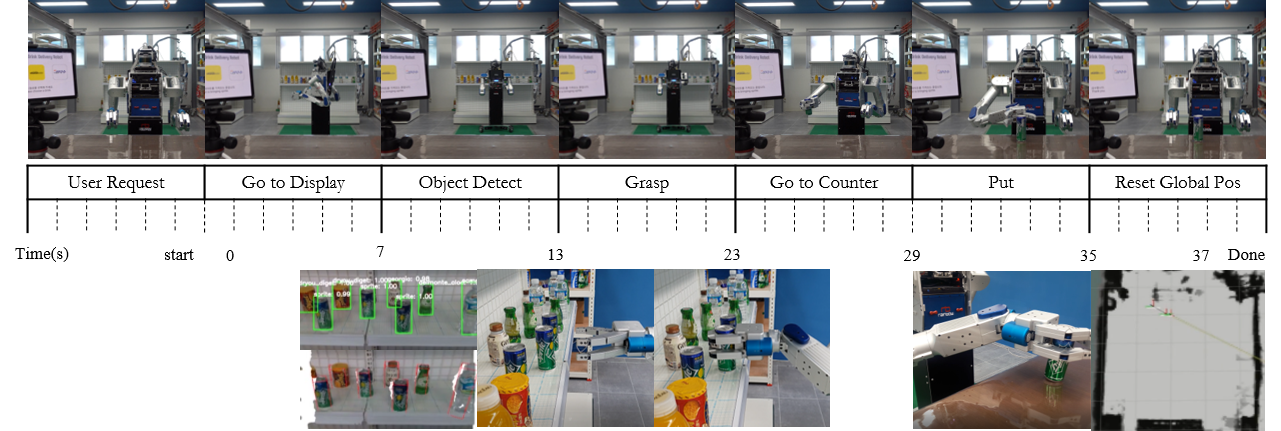}
    \caption{Time lapse and state machine transitions shown for robot butler serving 1 requested drink}
    \label{fig:time lapse}
\end{figure*}

For both methods of localization, we first create a map of the mock-environment then the localization module. Although the robot has a maximum base speed of $3.5$ km/h, it is reduced to $2.5$ km/h due to poor visual odometry especially during sudden rotational motions from the waist. 

Localization performance is evaluated via VICON motion capture cameras, to provide ground truth, recording at $100$ Hz with sub millimeter precision. For both localization methods, we compute the maximum and average deviation localization values against the ground truth with the following equations:

 \begin{equation} 
    \text{dev}_{\text{max}} =  \max (\sqrt{ (\hat{x}^2 - x^2) +   (\hat{y}^2 - y^2) } )
\end{equation}

 \begin{equation} 
    \text{dev}_{avg} =  \frac{1}{N}  \sum_{i=1}^{N} (\sqrt{ (\hat{x_i}^2 - x^2) +   (\hat{y_i}^2 - y^2) } 
\end{equation}
As expected, LiDAR-IMU based localization showed more accurate results, resulting in a maximum deviation of $0.13$ m and average deviation of $0.06$ m while RGB-D based localization resulted in $0.15$ m and $0.05$ m, respectively. As shown in Figure~\ref{fig:slam}, the performance between the two can be distinguished during the robots rotation motions. The inferior performance of RGB-D based localization is most likely caused due to motion blurring caused during rotations and therefore results in a reduced performance of feature detection for visual odometry. 
Although LiDAR localization performed accurately even at high velocities within the mock setup, its deviation error increased up to $0.3$ m and was unaffected by the reduction of the robots base velocity in the public exhibition. In the large open-space area ($90$ m $\times$ $120$ m) with few static features and a highly dynamic scene, RGB-D localization with reduced base velocity demonstrated higher robustness.

\section{EXPERIMENT AND RESULTS}

\begin{figure}[t]
    \centering
    \includegraphics[scale=0.7]{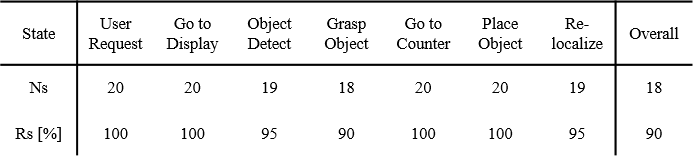}
    \caption{Success rate of robot's drink fetch in mock-environment by task state}
    \label{fig:result table mock}
\end{figure}

The  overall system performance of the robot fetching and serving drinks is evaluated in both an indoor real world setup and a public exhibition environment. We measure the total execution time per requested drink and success rate (Rs) for each state in the FSM. 
As shown in Figure~\ref{fig:time lapse} and Table~\ref{fig:result table mock}, from a total of $20$ trials (Ns) on average the entire service execution took $37$ seconds per drink with a success rate of $90\%$ in setup experiments. 

\begin{figure}[t]
    \centering
    \includegraphics[scale=0.7]{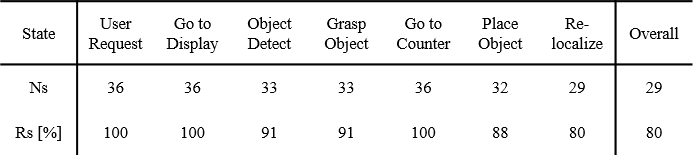}
    \caption{Success rate of robot's drink fetch in public exhibition by task state}
    \label{fig:result table ex}
\end{figure}

The lowest success rates occurred in the re-localize state, in which the robot base utilizes its localization data to reorient itself with respect to the global frame. Even at high velocity of $2.5$ km/h the robot was able to correctly re-localize itself in the mock-setup environment.
The Place Object task also had higher failure occurrences in the public exhibition in instances when the table for placement was significantly shifted from the audience members leaning, resulting in misaligned visual features from the initial map creation.   .

For comparison to human-like speed, we let humans participate in the fetch and serving drink task in the same proposed environment. On average, the entire service execution per drink resulted in $9$ seconds total across $6$ participants. 
Therefore our robot completing the task in $37$ seconds resulting in $24\%$ of the speed a human needs to fulfill the same task. This result is an improvement for service execution time where state-of-the-art service robots perform similar tasks in minutes, as discussed in Section II.  

Together with the robot butler service platform M-Hubo, we further publish a related video, code for each sub-module, and a detailed corresponding manual on the Hubo wikipage: \textit{www.KIrobotics.com.}

\section{CONCLUSIONS}

In this paper we presented a new robotic butler system for a wheeled humanoid that is capable of autonomously fetching requested objects in a static environment. This is achieved with our unique strategy of integrating a 3D object detection pipeline with a kinematically optimal manipulation planner to significantly increase execution speed at runtime.

The proposed system performed at 24\% of the speed a human needs to fulfill the same task. 
The system demonstrated high success rate of 90\% in our environment setup but reflected reduced performance of 80\% success rate in a more dynamic public exhibition due to environmental variations during run-time. 
For future work, the robot should incorporate localization scheme that is robust to variations in the environment and include additional states for failure recovery, which would allow for fast fetch execution even in dynamic environments. This step would require dynamic path planning and high-level task planner rather than a simple FSM fixed for a demo purpose. In addition, learning strategies can be utilized in the future to reduce failures, uncertainties, and unsafe states to increase success rate. Lastly, overall execution time can be further reduced even in dynamic environments by incorporating faster sampling motion planners described in this work~\cite{huboplan}.

\section{ACKNOWLEDGEMENT}
This work was supported by the Technology Innovation Program (or Industrial Strategic Technology Development Program (0070171, Development of core technology for advanced locomotion/manipulation based on high-speed/power robot platform and robot intelligence) funded By the Ministry of Trade, industry \& Energy(MI, Korea).

\addtolength{\textheight}{-12cm}   









\bibliographystyle{./IEEEtran} 
\bibliography{./egbib2}

\begin{thebibliography}{10}
\providecommand{\url}[1]{#1}
\csname url@rmstyle\endcsname
\providecommand{\newblock}{\relax}
\providecommand{\bibinfo}[2]{#2}
\providecommand\BIBentrySTDinterwordspacing{\spaceskip=0pt\relax}
\providecommand\BIBentryALTinterwordstretchfactor{4}
\providecommand\BIBentryALTinterwordspacing{\spaceskip=\fontdimen2\font plus
\BIBentryALTinterwordstretchfactor\fontdimen3\font minus
  \fontdimen4\font\relax}
\providecommand\BIBforeignlanguage[2]{{%
\expandafter\ifx\csname l@#1\endcsname\relax
\typeout{** WARNING: IEEEtran.bst: No hyphenation pattern has been}%
\typeout{** loaded for the language `#1'. Using the pattern for}%
\typeout{** the default language instead.}%
\else
\language=\csname l@#1\endcsname
\fi
#2}}

\bibitem{oldage}
``Age dependency ratio,'' \url{https://data.worldbank.org/indicator}.

\bibitem{careobot}
U.~Reiser, C.~Connette, J.~Fischer, J.~Kubacki, A.~Bubeck, F.~Weisshardt,
  J.~Theo, C.~Parlitz, M.~H¨agele, and A.~Verl, ``Care-o-bot®3 - creating a
  product vision for service robot applications by integrating design and
  technology,'' 2009.

\bibitem{herb}
S.~S. Srinivasa, D.~Ferguson, C.~J. Helfrich, D.~Berenson, A.~Collet,
  R.~Diankov, G.~Gallagher, G.~Hollinger, J.~Kuffner, and M.~V. Weghe, ``Herb:
  a home exploring robotic butler,'' 2010.

\bibitem{pr2smach}
J.~Bohren, R.~B. Rusu, E.~G. Jones, E.~Marder-Eppstein, C.~Pantofaru, M.~Wise,
  L.~Mosenlechner, W.~Meeussen, , and S.~Holzer, ``Towards autonomous robotic
  butlers: Lessons learned with the pr2,'' 2011.

\bibitem{rollinjustin}
B.~B¨auml, O.~Birbach, T.~Wimb¨ock, U.~Frese, A.~Dietrich, and G.~Hirzinger,
  ``Catching flying balls with a mobile humanoid: System overview and design
  considerations,'' 2011.

\bibitem{armar}
N.~Vahrenkamp, M.~Do, T.~Asfour, and R.~Dillmann, ``Integrated grasp and motion
  planning,'' 2006.

\bibitem{servicerobot}
Y.~Pyo, K.~Nakashimaa, S.~Kuwahataa, R.~Kurazumeb, T.~Tsuji, K.~Morookab, and
  T.~Hasegawac, ``Service robot system with an informationally structured
  environment,'' 2015.

\bibitem{jungDRC}
T.~Jung, J.~Lim, H.~Bae, K.~K. Lee, H.-M. Joe, and J.-H. Oh, ``Development of
  the humanoid disaster response platform drc-hubo+,'' 2018.

\bibitem{redmon2016yolo9000}
J.~Redmon and A.~Farhadi, ``Yolo9000: better, faster, stronger,'' \emph{{Proc.
  of Computer Vision and Pattern Recognition (CVPR)}}, 2017.

\bibitem{liu2016ssd}
W.~Liu, D.~Anguelov, D.~Erhan, C.~Szegedy, S.~Reed, C.-Y. Fu, and A.~C. Berg,
  ``Ssd: Single shot multibox detector,'' in \emph{{Proc. of European Conf. on
  Computer Vision (ECCV)}}.\hskip 1em plus 0.5em minus 0.4em\relax Springer,
  2016.

\bibitem{woo2017stairnet}
S.~Woo, S.~Hwang, and I.~S. Kweon, ``Stairnet: Top-down semantic aggregation
  for accurate one shot detection,'' in \emph{{Proc. of Winter Conference on
  Applications of Computer Vision (WACV)}}, 2018.

\bibitem{jang2018passd}
H.-D. Jang, S.~Woo, P.~Benz, J.~Park, and I.~S. Kweon, ``Propose-and-attend
  single shot detector,'' is in the progress of under review.

\bibitem{song2016deepsliding}
S.~Song and J.~Xiao, ``Deep sliding shapes for amodal 3d object detection in
  rgb-d images,'' in \emph{The IEEE Conference on Computer Vision and Pattern
  Recognition (CVPR)}, June 2016.

\bibitem{ren2015faster}
S.~Ren, K.~He, R.~Girshick, and J.~Sun, ``Faster r-cnn: Towards real-time
  object detection with region proposal networks,'' in \emph{{Proc. of Neural
  Information Processing Systems (NIPS)}}, 2015, pp. 91--99.

\bibitem{everingham2010pascal}
M.~Everingham, L.~Van~Gool, C.~K. Williams, J.~Winn, and A.~Zisserman, ``The
  pascal visual object classes (voc) challenge,'' in \emph{{Int'l Journal of
  Computer Vision (IJCV)}}.\hskip 1em plus 0.5em minus 0.4em\relax Springer,
  2010.

\bibitem{PRM}
H.~H~Gonzalez-Banos, D.~Hsu, and J.-C. Latombe, ``Motion planning: Recent
  developments,'' 2006.

\bibitem{moveit}
S.~Chitta, I.~Sucan, and S.~Cousins, ``Moveit! [ros topics],'' 2012.

\bibitem{equationpath}
D.~Youakim and P.~Ridao, ``Motion planning survey for autonomous mobile
  manipulators underwater manipulator case study.''\hskip 1em plus 0.5em minus
  0.4em\relax Robotics and Autonomous Systems 107, 2018.

\bibitem{equationsmooth}
M.~Zucker, N.~Ratliff, A.~Dragan, M.~Pivtoraiko, M.~Klingensmith, C.~Dellin,
  J.~A.~D. Bagnell, and S.~Srinivasa, ``Chomp: Covariant hamiltonian
  optimization for motion planning,'' 2013.

\bibitem{cartographer}
W.~Hess, D.~Kohler, H.~Rapp, and D.~Andor, ``Real-time loop closure in 2d lidar
  slam.''\hskip 1em plus 0.5em minus 0.4em\relax IEEE International Conference
  on Robotics and Automation, 2016.

\bibitem{rtabmap}
M.~Labbe and F.~Michaud, ``Appearance-based loop closure detection in real-time
  for large-scale and long-term operation.''\hskip 1em plus 0.5em minus
  0.4em\relax IEEE Transactions on Robotics, vol. 29, no. 3, pp. 734-745, 2013.

\bibitem{huboplan}
M.~Kang, D.~Kim, and S.-E. Yoon, ``Harmonious sampling for mobile manipulation
  planning.''\hskip 1em plus 0.5em minus 0.4em\relax IEEE/RSJ International
  Conference on Intelligent Robots and Systems, in press 2019.

\end{thebibliography}

\end{document}